# Adapted Approach for Fruit Disease Identification using Images

*Shiv Ram Dubey, GLA University Mathura, India*

*Anand Singh Jalal, GLA University Mathura, India*

## ABSTRACT

*Diseases in fruit cause devastating problem in economic losses and production in agricultural industry worldwide. In this paper, an adaptive approach for the identification of fruit diseases is proposed and experimentally validated. The image processing based proposed approach is composed of the following main steps; in the first step K-Means clustering technique is used for the defect segmentation, in the second step some state of the art features are extracted from the segmented image, and finally images are classified into one of the classes by using a Multi-class Support Vector Machine. We have considered diseases of apple as a test case and evaluated our approach for three types of apple diseases namely apple scab, apple blotch and apple rot. Our experimental results express that the proposed solution can significantly support accurate detection and automatic identification of fruit diseases. The classification accuracy for the proposed solution is achieved up to 93%.*

*Keywords: K-Means Clustering, Local Binary Pattern, Multi-class Support Vector Machine, Texture Classification*

## INTRODUCTION

Recognition system is a 'grand challenge' for the computer vision to achieve near human levels of recognition. In the agricultural sciences, images are the important source of data and information. To reproduce and report such data photography was the only method used in recent years. It is difficult to process or quantify the photographic data mathematically. Digital image analysis and image processing technology circumvent these problems based on the advances in computers and microelectronics associated with traditional photography. This tool helps to improve images from microscopic to telescopic visual range and offers a scope for their analysis.

Monitoring of health and detection of diseases is critical in fruits and trees for sustainable agriculture. To the best of our knowledge, no sensor is available commercially for the real time assessment of trees health conditions. Scouting is the most widely used method for monitoring stress in trees, but it is expensive, time-consuming and labor-intensive process. Polymerase chain reaction which is a molecular technique used for the identification of fruit diseases but it requires detailed sampling and processing.

The various types of diseases on fruits determine the quality, quantity, and stability of yield. The diseases in fruits not only reduce the yield but also deteriorate the variety and its withdrawal from the cultivation. Early detection of disease and crop health can facilitate the control of fruit diseases through proper management approaches such as vector control through fungicide applications, disease-specific chemical applications and pesticide applications; and improved productivity. The classical approach for detection and identification of fruit diseases is based on the naked eye observation by experts. In some of the developing countries, consultation with experts is a time consuming and costly affair due to the distant locations of their availability.

Fruit diseases can cause significant losses in yield and quality appeared in harvesting. For example, soybean rust (a fungal disease in soybeans) has caused a significant economic loss and just by removing 20% of the infection, the farmers may benefit with an approximately 11 million-dollar profit (Roberts et al., 2006). Some fruit diseases also infect other areas of the tree causing diseases of twigs, leaves and branches. An early detection of fruit diseases can aid in decreasing such losses and can stop further spread of diseases.

A lot of work has been done to automate the visual inspection of the fruits by machine vision with respect to size and color. However, detection of defects in the fruits using images is still problematic due to the natural variability of skin color in different types of fruits, high variance of defect types, and presence of stem/calyx. To know what control factors to consider next year to overcome similar losses, it is of great significance to analyze what is being observed.

The approach introduced in this paper can be used for designing automatic systems for agricultural process using images from distant farm fields. Several applications of image processing technology have been developed for the agricultural operations. These applications involve implementation of the camera based hardware systems or color scanners for inputting the images. We have attempted to extend image processing and analysis technology to a broad spectrum of problems in the field of agriculture. The computer based image processing is undergoing rapid evolution with ever changing computing systems. The dedicated imaging systems available in the market, where user can press a few keys and get the results, are not very versatile and more importantly, they have a high price tag on them. Additionally, it is hard to understand as to how the results are being produced.

Diseases appear as spots on the fruits and if not treated on time, cause severe losses. Excessive uses of pesticide for fruit disease treatment increases the danger of toxic residue level on agricultural products and has been identified as a major contributor to the ground water contamination. Pesticides are also among the highest components in the production cost thus their use must be minimized. Therefore, we have attempted to give an approach which can identify the diseases in the fruits as soon as they produce their symptoms on the growing fruits such that a proper management application can be applied.

Some common diseases of apple fruits are apple scab, apple rot, and apple blotch (Hartman, 2010). Apple scabs are gray or brown corky spots. Apple rot infections produce slightly sunken, circular brown or black spots that may be covered by a red halo. Apple blotch is a fungal disease and appears on the surface of the fruit as dark, irregular or lobed edges.

In this paper, we propose and experimentally evaluate an adaptive approach for the identification of fruit diseases using images. The proposed approach is composed of the following steps; in first step the fruit images are segmented using K-Means clustering technique, in second step, some state-of-the-art features are extracted from the segmented image, and finally, fruit diseases are classified using a Multi-class Support Vector Machine. We show the significance of using clustering technique for the disease segmentation and Multi-class Support Vector Machine as a classifier for the automatic classification of fruit diseases. In order to validate the proposed approach, we have considered three types of the diseases in apple; apple blotch, apple rot and apple scab. The experimental results shows that the proposed approach can significantly achieve accurate detection and automatic identification of fruit diseases.

## LITERATURE REVIEW

In this section, we focus on the previous work done by several researchers in the area of image categorization and fruit diseases identification. Fruit disease identification can be seen as an instance of image categorization.

Recently, a lot of activity in the area of image categorization has been done. Major works performing defect segmentation of fruits are done using simple threshold approach (Li, Wang, & Gu, 2002; Mehl et al., 2002). A globally adaptive threshold method (modified version of Otsu's approach) to segment fecal contamination defects on apples are presented by Kim et al. (2005). Classification-based methods attempt to partition pixels into different classes using different classification methods. Bayesian classification is the most used method by researchers Kleynen, Leemans, & Destain (2005) and Leemans, Magein, & Destain (1999), where pixels are compared with a pre-calculated model and classified as defected or healthy. Unsupervised classification does not benefit any guidance in the learning process due to lack of target values. This type of approach was used by Leemans, Magein, & Destain (1998) for defect segmentation.

The spectroscopic and imaging techniques are unique disease monitoring approaches that have been used to detect diseases and stress due to various factors, in plants and trees. Current research activities are towards the development of such technologies to create a practical tool for a large-scale real-time disease monitoring under field conditions.

Various spectroscopic and imaging techniques have been studied for the detection of symptomatic and asymptomatic plant and fruit diseases. Some the methods are: fluorescence imaging used by Bravo et al. (2004); multispectral or hyperspectral imaging used by Moshou et al. (2006); infrared spectroscopy used by Spinelli, Noferini, & Costa (2006); visible/multiband spectroscopy used by Yang, Cheng, & Chen (2007); Chen et al. (2008), and nuclear magnetic resonance (NMR) spectroscopy used by Choi et al. (2004). Hahn (2009) reviewed multiple methods (sensors and algorithms) for pathogen detection, with special emphasis on postharvest diseases. Several techniques for detecting plant diseases is reviewed by Sankarana et al. (2010) such as, Molecular techniques, Spectroscopic techniques (Fluorescence spectroscopy and Visible and infrared spectroscopy), and Imaging techniques (Fluorescence imaging and Hyper-spectral imaging).

A ground-based real-time remote sensing system for detecting diseases in arable crops under field conditions is developed by Moshou, (2005), which considers the early stage of disease development. The authors have used hyper-spectral reflection images of infected and healthy plants with an imaging spectrograph under ambient lighting conditions and field circumstances. They have also used multi-spectral fluorescence images simultaneously using UV-blue excitation on the same plants. They have shown that it is possible to detect presence of disease through the comparison of the 550 and 690 nm fluorescence images.

Large scale plantation of oil palm trees requires on-time detection of diseases as the ganoderma basal stem rot disease was present in more than 50% of the oil palm plantations in Peninsular Malaysia. To deal with this problem, airborne hyper-spectral imagery offers a better solution (Shafri & Hamdan, 2009) in order to detect and map the oil palm trees that were affected by the disease on time. Airborne hyper-spectral has provided data on user requirement and has the capability of acquiring data in narrow and contiguous spectral bands which makes it possible to discriminate between healthy and diseased plants better compared to multispectral imagery. Citrus canker is among the most devastating diseases that affect marketability of citrus crops. In (Qin et al., 2009), a hyper-spectral imaging approach is developed for detecting canker lesions on citrus fruit and hyper-spectral imaging system is developed for acquiring reflectance images from citrus samples in the spectral region from 450 to 930 nm. In (Purcell et al., 2009), the authors have investigated the power of NIR spectroscopy as an alternative to rate clones of sugarcane leaf spectra from direct measurement and examined its potential using a calibration model to successfully predict resistance ratings based on a chemometrics approach such as partial least squares. To populate the nature of the NIR sample, they have undertaken a scanning electron microscopy study of the leaf substrate.

Marcassa et al. (2006) have applied laser-induced fluorescence spectroscopy to investigate biological processes in orange trees. They have investigated water stress and Citrus Canker, which is a disease produced by the Xanthomonas axonopodis pv. citri bacteria. They have discriminated the Citrus Canker's contaminated leaves from the healthy leaves using a more complex analysis of the fluorescence spectra. However, they were unable to discriminate it from another disease.

## FRUIT DISEASE IDENTIFICATION

Image categorization, in general, relies on combinations of structural, statistical and spectral approaches. Structural approaches describe the appearance of the object using well-known primitives, for example, patches of important parts of the object. Statistical approaches represent the objects using local and global descriptors such as mean, variance, and entropy. Finally, spectral approaches use some spectral space representation to describe the objects such as Fourier spectrum (Gonzalez & Woods, 2007). In this paper, we introduce a method which exploits statistical color and texture descriptors to identify fruit diseases in a multi-class scenario.

The steps of the proposed approach are shown in the Figure 1. Defect segmentation, feature extraction, training and classification are the major tasks to be performed. For the fruit disease identification problem, precise image segmentation is required; otherwise the features of the non-infected region will dominate over the features of the infected region. K-means based defect segmentation is used to detect the region of interest which is the infected part only in the image. The proposed approach operates in two phases, i.e. training and classification. Training is required to learn the system with the characteristics of each type of diseases. First we extract the feature from the segmented portion of the images that are being used for the training and store in a feature database. Then, we train support vector machine with the features stored in the feature database. Finally any input image can be classified into one of the classes using feature derived from segmented part of the input image and trained support vector machine.

### Defect Segmentation

Image segmentation is a convenient and effective method for detecting foreground objects in images with stationary background. Background subtraction is a commonly used class of techniques for segmenting objects of interest in a scene. This task has been widely studied in the literature. Specular reflections, background clutter, shading and shadows are the major factors that affect the efficiency of the system. Therefore, in order to reduce the scene complexity, it might be interesting to perform image segmentation focusing on the object's description only.

K-means clustering technique is used for the defect segmentation. Images are partitioned into four clusters in which one or more cluster contains only infected region of the fruit. K-means clustering algorithm was developed by J. MacQueen (1967) and later by J. A. Hartigan & M. A. Wong (179). The K-means clustering algorithms classify the objects (pixels in our problem) into K number of classes based on a set of features. The classification is carried out by minimizing the sum of squares of distances between the data objects and the corresponding cluster.

*Figure 1. Proposed Approach for the Fruit Disease Identification*

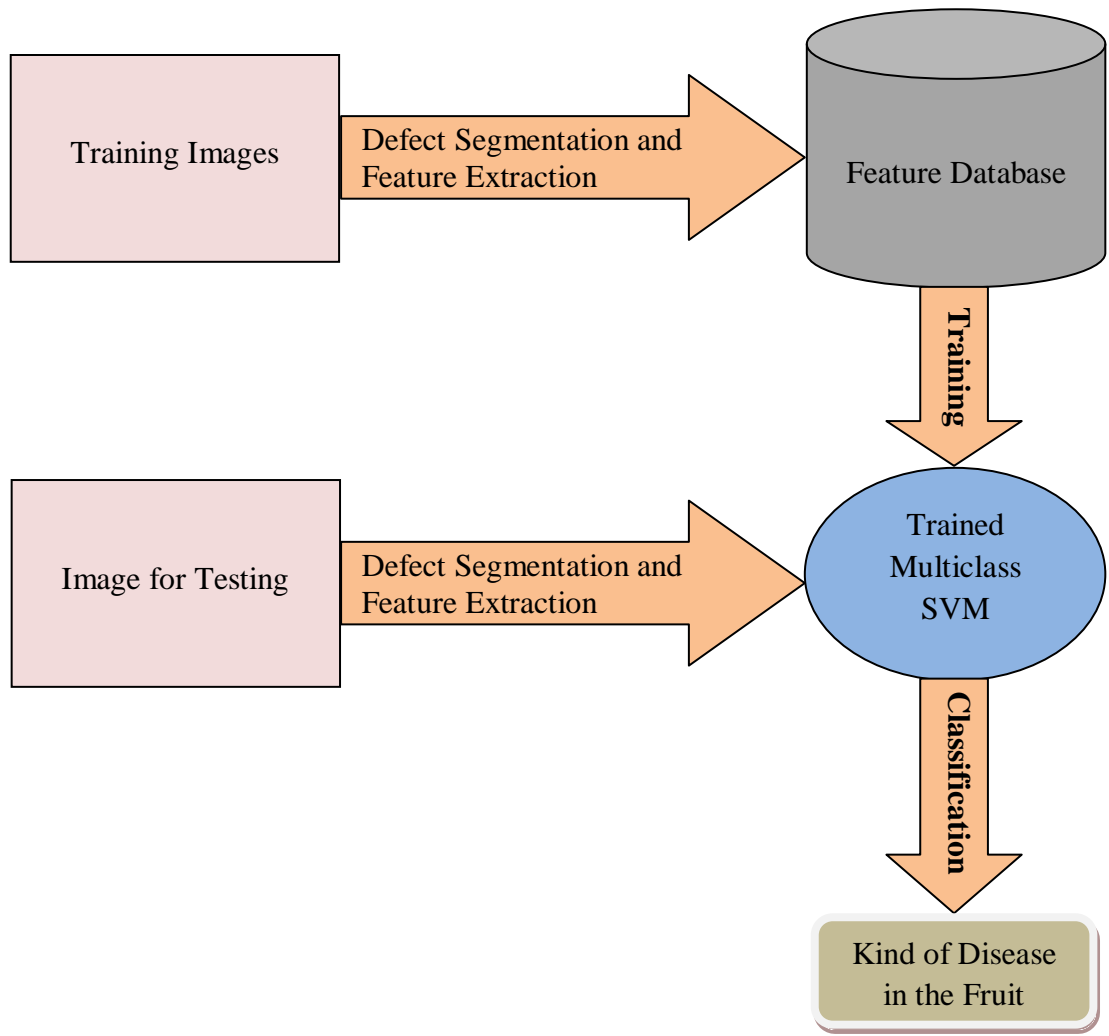

*K-Means Image Segmentation Algorithm*

Step 1. Read input image.

Step 2. Transform image from RGB to L*a*b* color space.

Step 3. Classify colors using K-Means clustering in 'a*b*' space.

Step 4. Label each pixel in the image from the results of K-Means.

Step 5. Generate images that segment the image by color.

Step 6. Select disease containing segment.

In this experiment, squared Euclidean distance is used for the K-means clustering. We use L*a*b* color space because the color information in the L*a*b* color space is stored in only two channels (i.e. a* and b* components), and it causes reduced processing time for the defect segmentation. In this experiment input images are partitioned into four segments. From the empirical observations it is found that using 3 or 4 cluster yields good segmentation results. Figure 2 demonstrates the output of K-Means clustering for an apple fruit infected with apple scab disease. Figure 3 also depicts some more defect segmentation results using the K-mean clustering technique.

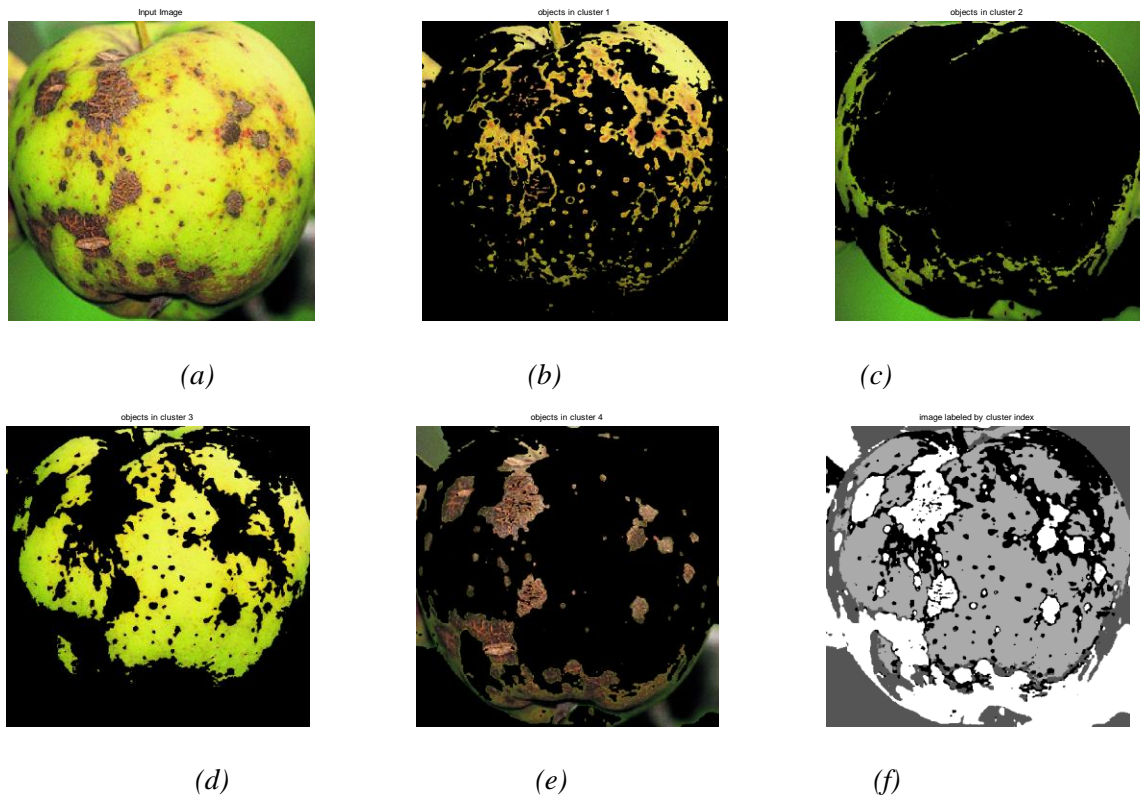

*Figure 2. K-Means clustering for an apple fruit that is infected with apple scab disease (a) The infected fruit image, (b) first cluster, (c) second cluster, (d) third cluster, and (e) fourth cluster, respectively, (f) single gray-scale image colored based on their cluster index.*

*Figure 3. Some defect segmentation results (a) Images before segmentation, (b) Images after segmentation*

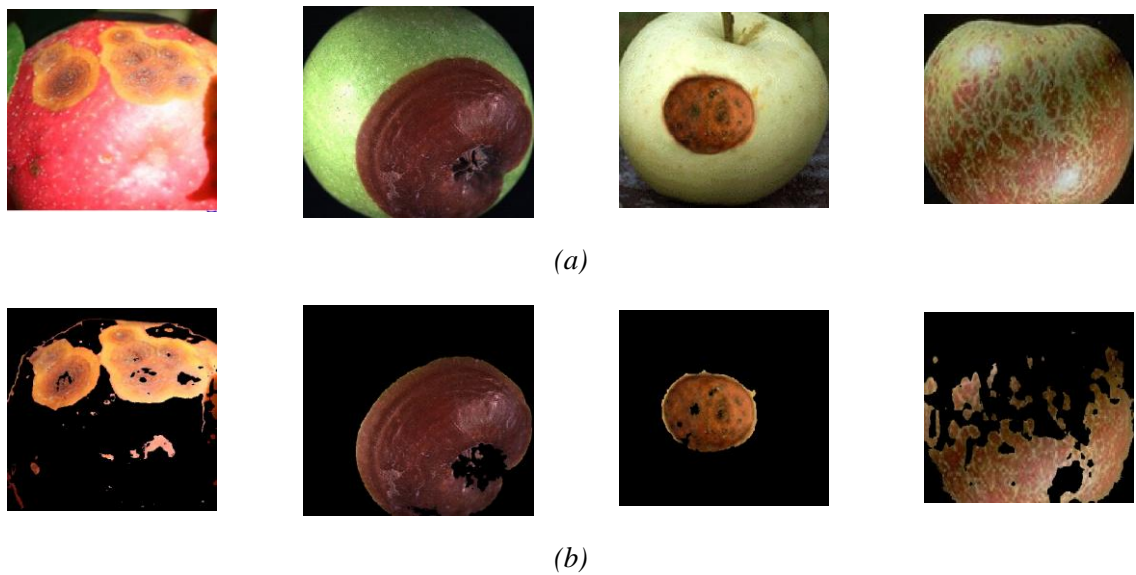

## Feature Extraction

We have used some state-of-the-art color and texture features to validate the accuracy and efficiency of the proposed approach. The features used for the fruit disease identification problem are Global Color Histogram, Color Coherence Vector, Local Binary Pattern, and Completed Local Binary Pattern.

### Global Color Histogram (GCH)

The Global Color Histogram (GCH) is the simplest approach to encode the information present in an image (Gonzalez & Woods, 2007). A GCH is a set of ordered values, for each distinct color, representing the probability of a pixel being of that color. Uniform normalization and quantization are used to avoid scaling bias and to reduce the number of distinct colors (Gonzalez & Woods, 2007).

### Color Coherence Vector (CCV)

An approach to compare images based on color coherence vectors are presented by Pass, Zabih, & Miller (1997). They define color coherence as the degree to which image pixels of that color are members of a large region with homogeneous color. These regions are referred as coherent regions. Coherent pixels are belongs to some sizable contiguous region, whereas incoherent pixels are not. In order to compute the CCVs, the method blurs and discretizes the image's color-space to eliminate small variations between neighboring pixels. Then, it finds the connected components in the image in order to classify the pixels of a given color bucket is either coherent or incoherent. After classifying the image pixels, CCV computes two color histograms: one for coherent pixels and another for incoherent pixels. The two histograms are stored as a single histogram.

### Local Binary Pattern (LBP)

Given a pixel in the input image, LBP is computed by comparing it with its neighbors (Ojala, Pietikainen, & Maenpaa, 2002):

$$LBP_{N,R} = \sum_{n=0}^{n-1} s(v_n - v_c)2^n, s(x) = \begin{cases} 1, x \geq 0 \\ 0, x < 0 \end{cases} \quad (1)$$

Where, $v_c$ is the value of the central pixel, $v_n$ is the value of its neighbors, $R$ is the radius of the neighborhood and $N$ is the total number of neighbors. Suppose the coordinate of $v_c$ is (0, 0), then the coordinates of $v_n$ are $(R\cos(2\pi n/N), R\sin(2\pi n/N))$. The values of neighbors that are not present in the image grids may be estimated by interpolation. Let the size of image is $I*J$. After the LBP code of each pixel is computed, a histogram is created to represent the texture image:

$$H(k) = \sum_{i=1}^{I}\sum_{j=1}^{J} f(LBP_{N,R}(i,j),k), k \in [0,K],$$
$$f(x,y) = \begin{cases} 1, x = y \\ 0, otherwise \end{cases} \quad (2)$$

Where, $K$ is the maximal LBP code value.

In this experiment the value of '$N$' and '$R$' are set to '8' and '1' respectively to compute the LBP feature.

### Completed Local Binary Pattern (CLBP)

LBP feature considers only signs of local differences (i.e. difference of each pixel with its neighbors) whereas CLBP feature considers both signs (S) and magnitude (M) of local differences as well as original center gray level (C) value (Guo, Zhang, & Zhang, 2010). CLBP feature is the combination of three features, namely CLBP_S, CLBP_M, and CLBP_C. CLBP_S is the same as the original LBP and used to code the sign information of local differences. CLBP_M is used to code the magnitude information of local differences:

$$CLBP_{N,R} = \sum_{n=0}^{n-1} t(m_n, c)2^n, t(x,c) = \begin{cases} 1, x \geq c \\ 0, x < c \end{cases} \quad (3)$$

Where, $c$ is a threshold and set to the mean value of the input image in this experiment.

CLBP_C is used to code the information of original center gray level value:

$$CLBP_{N,R} = t(g_c, c_I), t(x,c) = \begin{cases} 1, x \geq c \\ 0, x < c \end{cases} \quad (4)$$

Where, threshold $c_I$ is set to the average gray level of the input image. In this experiment the value of '$N$' and '$R$' are set to '8' and '1' respectively to compute the CLBP feature.

## Supervised Learning and Classification

Supervised learning is a machine learning approach that aims to estimate a classification function f from a training data set. The trivial output of the function f is a label (class indicator) of the input object under analysis. The learning task is to predict the function outcome of any valid input object after having seen a sufficient number of training examples.

In the literature, there are many different approaches for supervised learning such as Linear Discriminant Analysis (LDA), Support Vector Machines (SVMs), Classification Trees, Neural Networks (NNs), and Ensembles of Classifiers (Bishop, 2006).

Recently, a unified approach is presented by Rocha et al. (2010) that can combine many features and classifiers. The author approaches the multi-class classification problem as a set of binary classification problem in such a way one can assemble together diverse features and classifier approaches custom-tailored to parts of the problem. They define a class binarization as a mapping of a multi-class problem onto two-class problems (divide-and-conquer) and referred binary classifier as a base learner. For N-class problem $N \times (N-1)/2$ binary classifiers will be needed where $N$ is the number of different classes.

*Table 1. Unique ID of each class*

|   | $x \times y$ | $x \times z$ | $y \times z$ |
|---|---|---|---|
| $x$ | +1 | +1 | 0 |
| $y$ | -1 | 0 | +1 |
| $z$ | 0 | -1 | -1 |

According to the author, the $ij^{th}$ binary classifier uses the patterns of class $i$ as positive and the patterns of class $j$ as negative. They calculate the minimum distance of the generated vector (binary outcomes) to the binary pattern (ID) representing each class, in order to find the final outcome. They have categorized the test case into a class for which distance between ID of that class and binary outcomes is minimum.

This approach can be understood by a simple three class problem. Let three classes are $x$, $y$, and $z$. Three binary classifiers consisting of two classes each (i.e., $x \times y$, $x \times z$, and $y \times z$) are used as base learners, and each binary classifier is trained with training images. Each class receives a unique ID as shown in Table 1. To populate the table is straightforward. First, we perform the binary comparison $x \times y$ and tag the class $x$ with the outcome +1, the class $y$ with −1 and set the remaining entries in that column to 0. Thereafter, we repeat the procedure comparing $x \times z$, tag the class $x$ with +1, the class $z$ with −1, and the remaining entries in that column with 0. In the last, we repeat this procedure for binary classifier $y \times z$, and tag the class $y$ with +1, the class $z$ with -1, and set the remaining entries with 0 in that column, where the entry 0 means a "Don't care" value. Finally, each row represents unique ID of that class (e.g., $y = [−1, +1, 0]$).

Each binary classifier results a binary response for any input example. Let's say if the outcomes for the binary classifier $x \times y$, $x \times z$, and $y \times z$ are +1, -1, and +1 respectively, then the input example belongs to that class which have the minimum distance from the vector [+1, -1, +1]. So the final answer is given by the minimum distance of

$$\min \ \text{dist}\big(\{+1,-1,+1\},(\{+1,+1,0\},\{-1,0,+1\},\{0,-1,-1\})\big)$$

In this experiment, we have used Multi-class Support Vector Machine (MSVM) as a set of binary Support Vector Machines (SVMs) for the training and classification.

## RESULTS AND DISCUSSIONS

In this section, first we discuss about the data set of apple fruit diseases and after present a detailed result of the fruit disease identification problem and discuss various issues regarding the performance and efficiency of the system. We consider two color-spaces (i.e. RGB and HSV color-space) and compare the performance of the system under these color spaces.

*Figure 4. Sample images from the data set of type (a) apple scab, (b) apple rot, (c) apple blotch, and (d) normal apple*

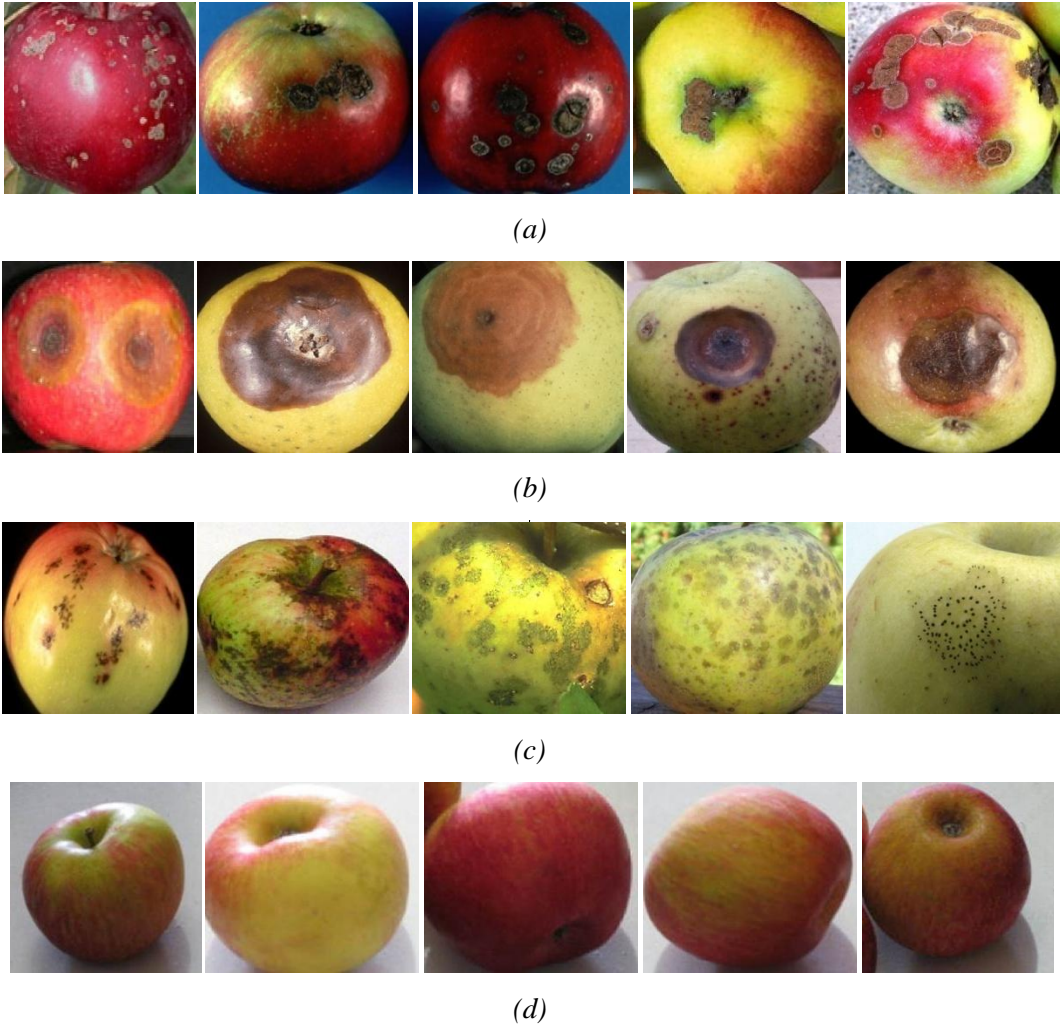

*(a)*

*(b)*

*(c)*

*(d)*

### Data Set Preparation

To demonstrate the performance of the proposed approach, we have used a data set of normal and diseased apple fruits, which comprises four different categories: Apple Blotch (104), Apple rot (107), Apple scab (100), and Normal Apple (80). The total number of apple fruit images (N) is 391. Figure 4 depicts the classes of the data set. Presence of a lot of variations in the type and color of apple makes the data set more realistic.

### Result Discussion

In the quest for finding the best categorization procedure and feature to produce classification, we have analyzed some color and texture based image descriptors derived from RGB and HSV stored images considering Multiclass Support Vector Machine (MSVM) as classifier. If we use M images per class for training then remaining N-4*M are used for testing. The accuracy of the proposed approach is defined as,

$$\text{Accuracy}(\%) = \frac{\text{Total number of images correctly classified}}{\text{Total number of images used for testing}} *100$$

Figure 5(a) and 5(b) shows the results for different features in the RGB and HSV color spaces respectively. The x-axis represents the number of images per class in the training set and the y-axis represents the accuracy for the test images.

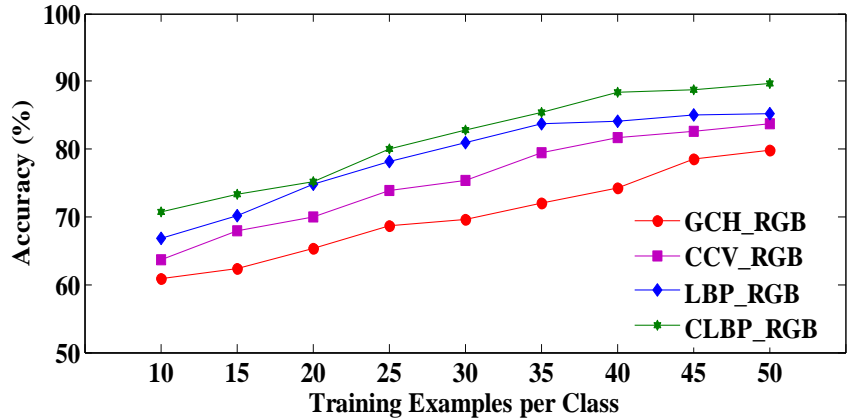

*(a) Using RGB color image*

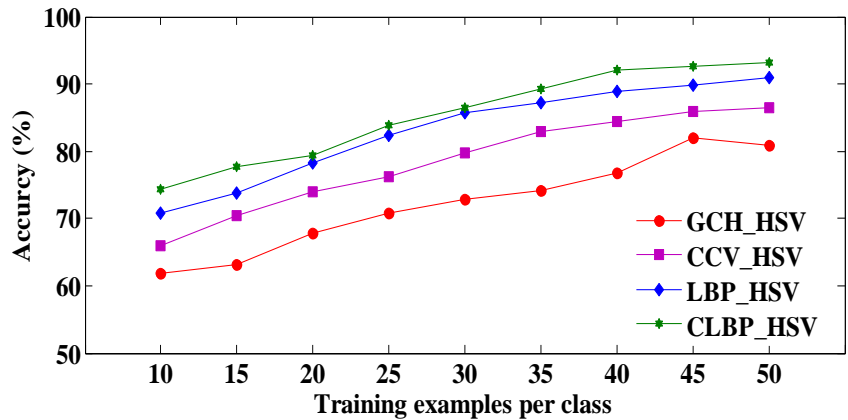

*(b) Using HSV color image*

*Figure 5. Accuracy (%) for the GCH, CCV, LBP, and CLBP features derived from RGB and HSV color images considering MSVM classifier.*

This experiment shows that GCH does not perform well and reported accuracy is lowest for it in both the color spaces. One possible explanation is that, GCH feature have only color information, it does not considers neighboring information. GCH uses simply frequency of each color, however CCV uses frequency of each color in coherent and incoherent regions separately and so it performs better than GCH in both color spaces.

From the Figure 5, it is clear that LBP and CLBP features yield better result than GCH and CCV features because both LBP and CLBP uses the neighboring information of each pixel in the image. Both LBP and CLBP are robust to illumination differences and they are more efficient in pattern matching because they use local differences which are computationally more efficient. In HSV color space with 50 training examples per class, the reported classification accuracy is 80.94% for GCH, 86.47% for CCV, 90.97% for LBP, and 93.14% for CLBP feature.

The LBP feature uses only the sign information of the local differences, even then, LBP reasonably represent the image local features because sign component preserves the major information of local differences. The CLBP feature exhibits more accurate result than LBP feature because CLBP feature uses both sign and magnitude component of local differences with original center pixel value.

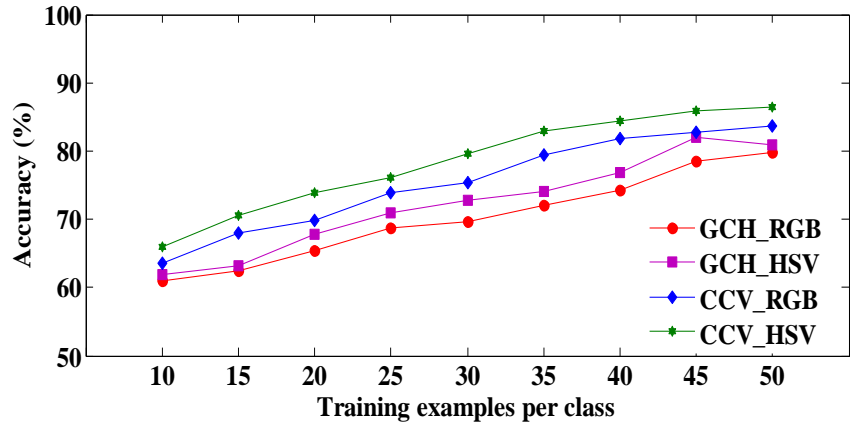

*(a) GCH and CCV feature*

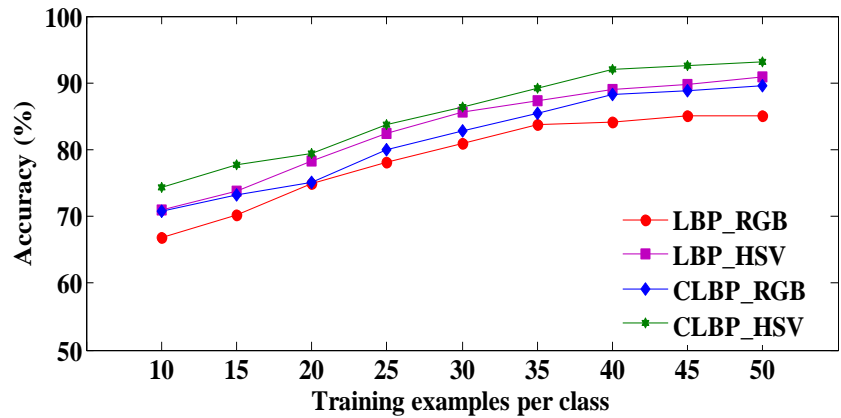

*(b) LBP and CLBP feature*

*Figure 6. Comparison of the accuracy achieved in RGB and HSV color space for the GCH, CCV, LBP, and CLBP features considering MSVM classifier.*

We also observe across the plots that each feature performs better in the HSV color space than the RGB color space as shown in the Figure 6 (a-b). For 45 training examples and CLBP feature, for instance, reported classification error is 88.74% in RGB and 92.65% in HSV.

One important aspect when dealing with apple fruit disease classification is the accuracy per class. This information points out the classes that need more attention when solving the confusions. Figure 7 and 8 depicts the accuracy for each one of 4 classes using LBP and CLBP features in RGB and HSV color spaces. Clearly, Apple Blotch is one class that needs attention in both color spaces. It yields the lowest accuracy when compared to other classes in both color spaces. Figure 7 and 8 also shows that, the behavior of Apple Rot is nearly same in each scenario.

Normal Apples are very easily distinguishable with diseased apples and a very good classification result is achieved for the Normal Apples in both color spaces as shown in Figure 7 and 8. For CLBP feature and HSV color space, for instance, reported classification accuracy are 89.88%, 90.71%, 96.66%, and 99.33% for the Apple Blotch, Apple Rot, Apple Scab, and Normal Apple respectively, resulting average accuracy 93.14% when training is done with 50 images per class.

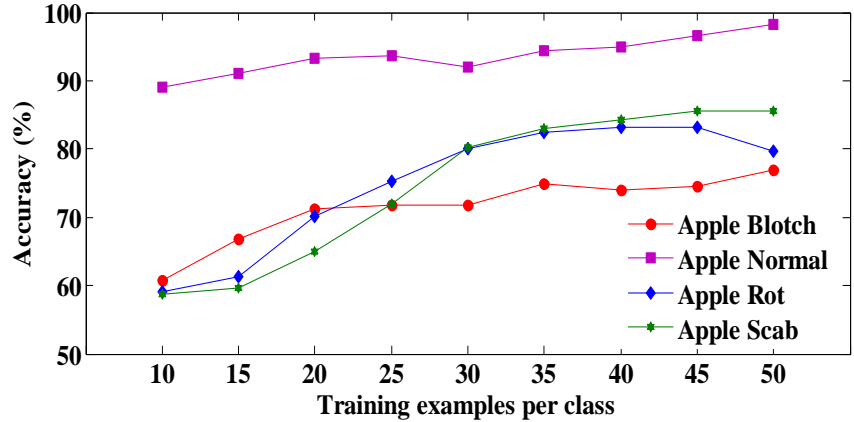

*Figure 7. Accuracy per class for the LBP features in RGB and HSV color spaces using MSVM as a classifier*

*(a) LBP in RGB color space*

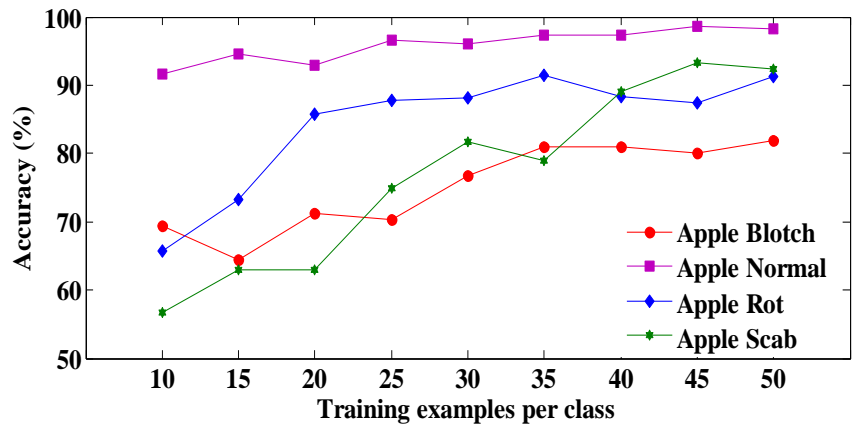

*(b) LBP in HSV color space*

## CONCLUSIONS

An image processing based approach is proposed and evaluated in this paper for fruit disease identification problem. The proposed approach is composed of mainly three steps. In the first step defect segmentation is performed using K-means clustering technique. In the second step features are extracted. In the third step training and classification are performed on a Multiclass SVM. We have used three types of apple diseases namely: Apple Blotch, Apple Rot, and Apple Scab as a case study and evaluated our program.

   Our experimental results indicate that the proposed solution can significantly support automatic detection and classification of apple fruit diseases. Based on our experiments, we have found that normal apples are easily distinguishable with the diseased apples and CLBP feature shows more accurate result for the identification of apple fruit diseases and achieved more than 93% classification accuracy. Further work includes consideration of fusion of more than one feature to improve the output of the proposed method.

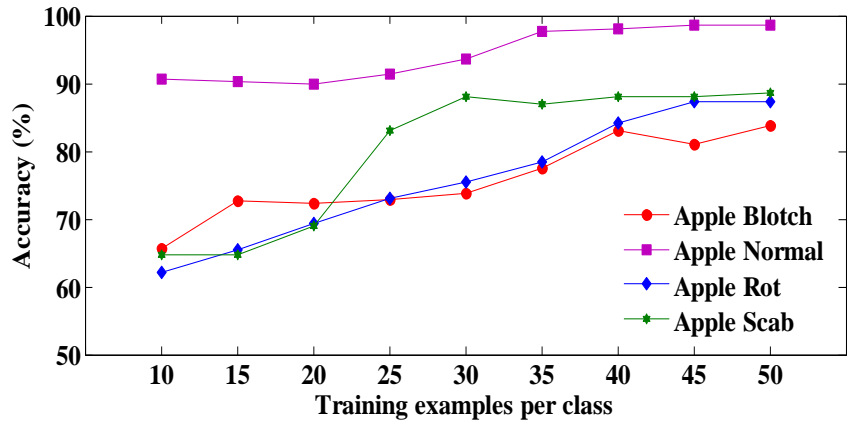

*Figure 8. Accuracy per class for the CLBP features in RGB and HSV color spaces using MSVM as a classifier*

*(a) CLBP in RGB color space*

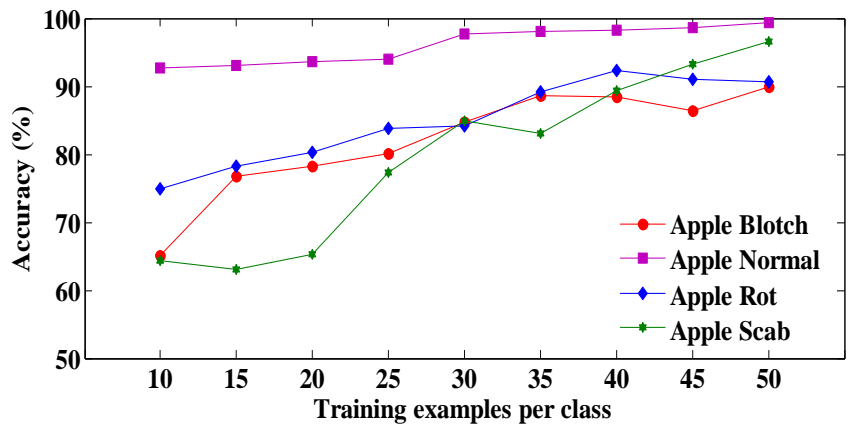

*(b) CLBP in HSV color space*